\documentclass[conference]{IEEEtran}
\IEEEoverridecommandlockouts
\usepackage{cite}
\usepackage{amsmath,amssymb,amsfonts}
\usepackage{algorithmic}
\usepackage{graphicx}
\usepackage{textcomp}
\usepackage{xcolor}
\usepackage{booktabs}
\def\BibTeX{{\rm B\kern-.05em{\sc i\kern-.025em b}\kern-.08em
    T\kern-.1667em\lower.7ex\hbox{E}\kern-.125emX}}
\begin{document}

\title{Touchless Intraoperative Image Access System Based on Vision-Based Hand Tracking
\thanks{$^{1}$This author is also affiliated with Fondazione IRCCS Istituto Neurologico Carlo Besta}
}
\author{
\IEEEauthorblockN{1\textsuperscript{st} Yin Lin}
\IEEEauthorblockA{DEIB\\
Polytechnic University of Milan\\
Milan, Italy\\
yin.lin@polimi.it}

\and
\IEEEauthorblockN{2\textsuperscript{nd} Domenico Aquino}
\IEEEauthorblockA{Neuroradiology Unit\\
Fondazione IRCCS Istituto Neurologico \\Carlo Besta\\
Milan, Italy\\
domenico.aquino@istituto-besta.it}

\and
\IEEEauthorblockN{3\textsuperscript{rd} Alberto Redaelli}
\IEEEauthorblockA{DEIB\\
Polytechnic University of Milan\\
Milan, Italy\\
alberto.redaelli@polimi.it}

\and
\IEEEauthorblockN{4\textsuperscript{th} Massimiliano Del Bene}
\IEEEauthorblockA{Neurosurgery 1\\
Fondazione IRCCS Istituto Neurologico \\Carlo Besta\\
Milan, Italy\\
massimiliano.delbene@istituto-besta.it}

\and
\IEEEauthorblockN{5\textsuperscript{th} Riccardo Barbieri}
\IEEEauthorblockA{DEIB\\
Polytechnic University of Milan\\
Milan, Italy\\
riccardo.barbieri@polimi.it}

\and
\IEEEauthorblockN{6\textsuperscript{th} Simona Ferrante$^{1}$}
\IEEEauthorblockA{DEIB\\
Polytechnic University of Milan\\
Milan, Italy\\
simona.ferrante@polimi.it}
}

\maketitle

\begin{abstract}
Touchless interaction with medical images is becoming increasingly important in the surgical field, where sterility and continuity of the operational workflow are essential requirements. This work presents a vision-based system for intraoperative navigation of medical images through hand gestures acquired using a single RGB camera. Unlike many existing solutions, the system does not require additional hardware or user-specific training. Hand tracking is performed in real time using MediaPipe Hands, which provides a 2.5D estimation of hand landmarks. Simple and intuitive gestures are then mapped into translation, rotation, and zoom commands, enabling continuous and natural interaction with the image viewer. The system architecture is independent from the visualization software and, for implementation simplicity, in this study it was integrated with PyVista. Performance was evaluated through frame-level logging and quantitative analysis of latency, stability, and interaction robustness metrics. Experimental results highlight real-time behavior, with reduced latencies and stable control, in line with the requirements of fluid interaction. The system demonstrates the feasibility of a low-cost touchless solution for intraoperative access to medical images, laying the groundwork for future clinical evaluations.
\end{abstract}


\section{Introduction}
Sterility has always been a fundamental requirement in surgical practice, as it is closely associated with the reduction of surgical site infections and the improvement of clinical outcomes \cite{I1}. Numerous studies have demonstrated that cross-contamination resulting from contact with non-sterile surfaces, such as keyboards, mice, and touchscreens, can increase the risk of intra- and postoperative infections \cite{I2}. This issue has become even more relevant with the emergence of the COVID-19 pandemic, which has further highlighted the need to limit physical contact between healthcare professionals and devices present in the operating room \cite{I3,I4}. In this context, numerous touchless software solutions have been proposed in the healthcare domain, with the aim of reducing direct interactions with the clinical environment. For example, O’Hara et al. \cite{I5} analyzed the use of contactless interfaces to access clinical data. Similarly, Mentis et al. \cite{I6} introduced touchless systems for the consultation of radiological images. More specifically in the context of the operating room, many studies have explored gesture-based interaction systems for intraoperative access to computed tomography, magnetic resonance imaging, and ultrasound images. Wachs et al. \cite{I7} presented one of the first gesture-based control systems to navigate medical images in the operating room. Subsequently, Jacob et al. \cite{I8} and several studies \cite{I10,I9} confirmed that the use of contactless interfaces allows surgeons to access intraoperative images without interrupting the procedure or compromising the sterile field.

In the present work, we focus on Non-Contact Intraoperative Image Access Systems (NCIAS). Such systems can be implemented through different approaches based on automatic recognition of hand gestures. The first approach is based on contact-free sensor-based systems, which require the installation of specific sensors or hardware in the operating room environment \cite{I11}. Meanwhile, the largest research stream uses active techniques. For example, studies \cite{I12,I13,I14,I15} employed Microsoft Kinect (MK), which takes advantage of depth sensors to improve the robustness of gesture recognition. Other studies \cite{I16,I17,I18}, instead, explored the use of the Leap Motion Controller (LMC). Kumar et al. \cite{I19,I20} combine MK and LMC to increase performance.
Despite the progress achieved, existing solutions present several limitations. Firstly, many systems are not sufficiently robust to input variability, due to inter-user differences, gesture execution speed, and environmental conditions \cite{I8,I12}. Secondly, the automatic recognition of the beginning and the end of a gesture remains an open challenge, particularly in continuous interaction scenarios \cite{I21}. Moreover, several approaches suffer from marked signer-dependency, requiring a user training period; for this reason, predefined gesture dictionaries have been proposed to standardize interaction and improve usability \cite{I7,I21}. Finally, many systems are environment-dependent, requiring accurate positioning of sensors or controllers, which limits flexibility and clinical adoption \cite{I12,I16}.

With the emergence of deep learning techniques \cite{I16a,I16b}, numerous AI-based models have been developed for the automatic recognition of hands and gestures, demonstrating significantly superior performance compared to traditional approaches. Tompson et al. \cite{I22} introduced one of the first models based on Convolutional Neural Networks (CNN) for hand pose estimation from depth images. Subsequently, Oberweger et al. \cite{I23} proposed a deep learning approach for the direct prediction of hand joint coordinates, showing greater robustness to pose variations and occlusions. In the RGB context, Simon et al. \cite{I24} and Zimmermann and Brox \cite{I25} demonstrated the possibility of accurately estimating the 3D hand pose from a single RGB image. For dynamic gesture recognition, Molchanov et al. \cite{I26} introduced architectures that combine three-dimensional CNNs and recurrent models to capture temporal information. More recently, Zhang et al. \cite{I27} presented MediaPipe Hands, a deep learning–based framework capable of providing real-time tracking of 21 hand landmarks using a single RGB camera.

This work represents, to the best of our knowledge, one of the first implementations of MediaPipe Hands \cite{I27} for touchless intraoperative image access. While gesture-based systems in sterile surgical environments have been previously proposed, our approach specifically leverages MediaPipe for real-time interaction with intraoperative imaging. The proposed system is designed to be robust to input variability, usable without a dedicated training period, such that intuitive for operators already familiar with touch interfaces. The recognition of the beginning, the end, and the transitions between gestures occurs in a smooth and natural manner, enabling continuous interaction. Moreover, the system is environment-independent, requiring only a standard camera, without the need for dedicated sensors or complex installations. 

\section{METHODS}

\subsection{Experimental Setup and Implementation}
All experiments were conducted on a laptop computer (MSI Cyborg 15 A13VE-603IT) equipped with an Intel Core i5-13420H processor, an RTX 4050 GPU, and an HD webcam with a resolution of 1280×720 pixels at 30 fps. The system relies exclusively on monocular RGB video acquisition and does not require depth sensors, wearable devices, or dedicated tracking hardware. Experiments were performed by three subjects in a closed indoor environment under natural lighting conditions. Subjects performed the gestures in a seated position, and the 10 interaction sessions were distributed across the three participants. The software was developed in Python (version 3.12.6) using MediaPipe \cite{I27} for real-time hand tracking and PyVista \cite{I28} for interactive three-dimensional visualization. Rendering is handled through a non-blocking Qt backend (BackgroundPlotter), enabling concurrent gesture processing and scene updates without interrupting the main execution loop.

\subsection{Hand Tracking and Gesture-Based Interaction}
Hand tracking is performed using MediaPipe Hands which returns 21 2.5D landmarks for each frame. Each landmark $l_{k}$ is represented by coordinates $x_{k},y_{k},z_{k}$ while the fingertip coordinates $x^{p}_{k},y^{p}_{k}$ are obtained by scaling the normalized coordinates according to the image resolution (formula \ref{f1}).
\begin{equation}
\label{f1}
    x^{p}_{k} = x_{k} \cdot W, \quad y^{p}_{k} = y_{k} \cdot H
\end{equation}
Where \( x_k, y_k \in [0,1] \) are the normalized image coordinates provided by MediaPipe, and W and H respectively indicate the width and height of the camera's images in pixels.

As a first demo, only three interaction modes were implemented, namely translation, indicated as SHIFT mode, rotation (ROTATE), and zoom (ZOOM). Mode selection is performed based on a combination of conditions on hand posture and depth dominance among the fingers. The shift mode is activated when the index finger is extended with a relative depth that exceeds a predefined threshold and is greater than that of the middle finger by a delta margin. The rotate mode is defined in an orthogonal manner, where the middle finger is used as the dominant finger. For both modes, the frame-to-frame increments are computed as:
\begin{equation}
    \Delta x_{k}^{p}(t) = x_{k}^{p}(t)-x_{k}^{p}(t-1), \quad \Delta y_{k}^{p}(t) = y_{k}^{p}(t)-y_{k}^{p}(t-1)
\end{equation}
Finally, the zoom mode is activated through a pinch gesture between the index finger and the thumb. The pinch distance is computed as the Euclidean distance between the fingertips in image space:
\begin{equation}
    d_{pinch}=\sqrt{(x^{p}_{index}-x^{p}_{thumb})^{2}+(y^{p}_{index}-y^{p}_{thumb})^{2}}
\end{equation}
and the zoom command is generated from the temporal variation of this distance.
\begin{equation}
    \Delta d_{pinch}(t) = d_{pinch}(t)-d_{pinch}(t-1)
\end{equation}
Where positive values of \( \Delta d_{pinch} \) correspond to a zoom-out operation (increasing finger distance), and negative values correspond to zoom-in.

To improve robustness, the zoom mode requires the ring and little fingers to be closed and is immediately deactivated when the middle finger is extended. When the mode is active, commands are generated continuously from the frame-to-frame displacement of the fingertips. To address input variability due to environment, operator, and hardware, all thresholds and gains are scaled through a single parameter defined as global sensitivity.

\begin{figure}
    \centering
    \includegraphics[width=1\linewidth]{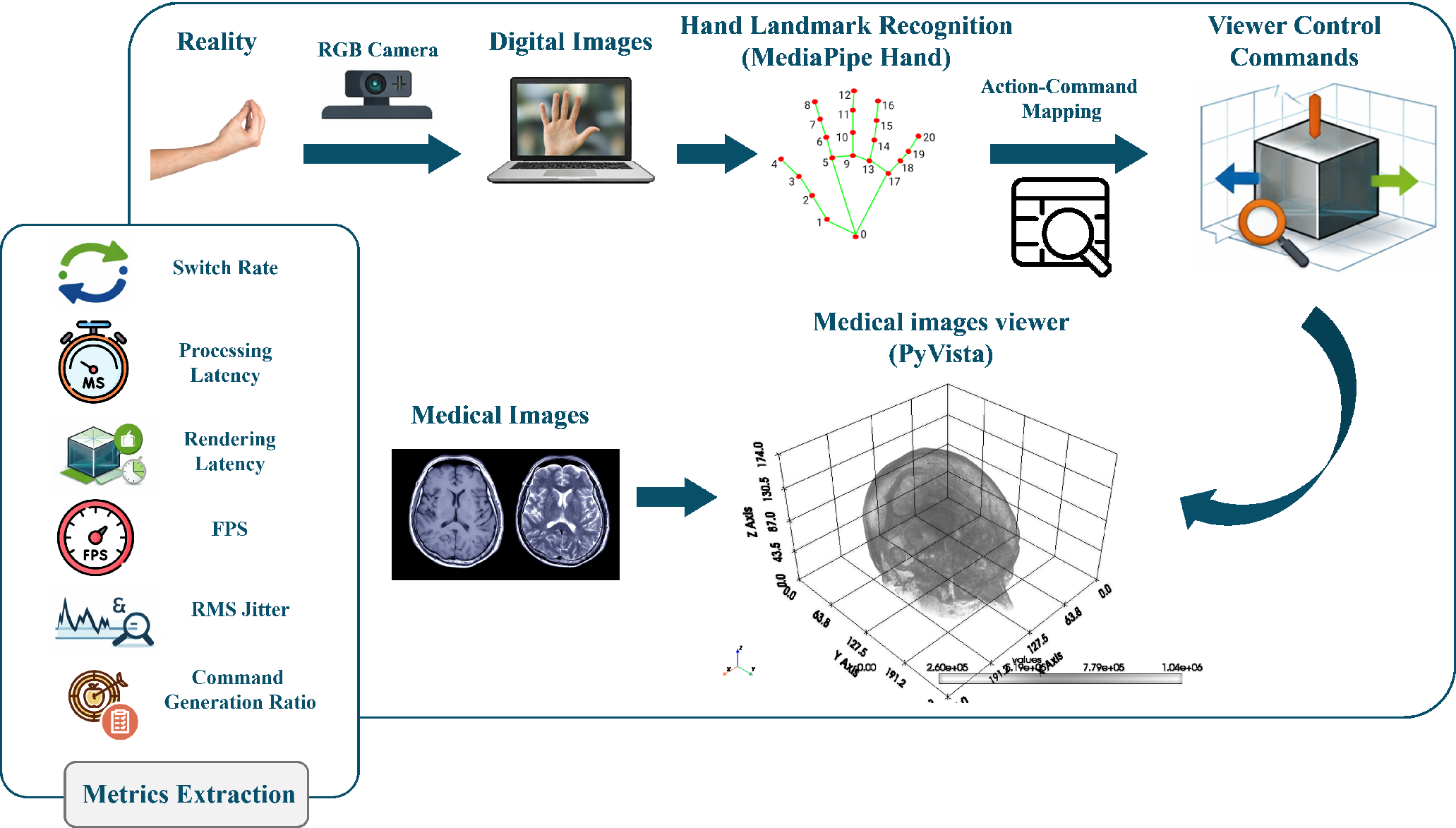}
    \caption{Overview of the gesture-based interaction pipeline.}
    \label{pipeline}
\end{figure}

The complete processing pipeline, from RGB image acquisition to gesture interpretation, command generation, and performance metrics extraction, is summarized in Fig.~\ref{pipeline}.

\subsection{Performance Logging and Evaluation Metrics}
A total of 10 interaction sessions were performed. During each session, frame-level data were recorded whenever a control mode was active. For each frame at time $t$, a record $r(t)$ was generated, with all records across sessions aggregated into a single CSV file. Each record contains the recognized mode, the recorded action, the fingertip positions in pixel coordinates, the relative depth, the pinch distance, and processing and rendering times. The recorded data are subsequently analyzed to derive quantitative metrics of stability and real-time performance. The evaluation is performed both for individual mode $m$ and from a global perspective.
Five metrics are used to evaluate each individual mode and six metrics are used for the global evaluation. We first define $T_{m} = \{t| \quad r(t)=m\}$ as the set of frames associated with mode m, with cardinality  $N_{m} = |T_{m}|$.
The first metric considered is the command-generation ratio (CMD-Gen ratio), which quantifies the probability that a gesture effectively induces a command. This index is defined as follows:
\begin{equation}
    R_{m} = \frac{1}{N_{m}}\sum_{t\in T_{m}}^{}a(t)
\end{equation}
with $a_{t}$ is equal to 1 if a command is issued at frame $t$ (i.e. a non-empty action is generated), and 0 otherwise.
The second metric is the RMS jitter (RMS jitter)expressed in percentage, which quantifies spatial sensibility during active interaction. For SHIFT and ROTATE modes, the index is computed as the square root of the mean squared frame-to-frame displacement divided by the image diagonal making the metric dimensionless and independent of image resolution:
\begin{equation}
    \begin{array}{l}
        J_{m} =\frac{1}{\sqrt{W^{2}+H^{2}}} \sqrt{\frac{1}{N_{m}}\sum_{t\in T_{m}}^{}\Delta x^{p}(t)^{2}+\Delta y^{p}(t)^{2}},  \\
        m\in \{SHIFT, ROTATE\} 
    \end{array}
\end{equation}
For the ZOOM mode, in which control is one-dimensional, the RMS jitter is defined based on the variation of the pinch distance:
\begin{equation}
    J_{ZOOM} = \frac{1}{\sqrt{W^{2}+H^{2}}}\sqrt{\frac{1}{N_{ZOOM}}\sum_{t\in T_{ZOOM}}^{}\Delta d(t)_{pinch}^{2}}
\end{equation}
The third metric is the average processing latency (Proc latency), which corresponds to the time required to acquire an input frame, estimate the hand landmarks, and infer both the active interaction mode and the associated command.  Based on this latency, an effective frame rate (FPS) can be derived as the ratio between 1000 and the average processing latency.
The fifth metric is the average rendering latency (Render latency), which measures the time needed to update the three-dimensional visualization after a command has been applied.
Finally, from a global perspective, the mode switching rate (Switching rate) is introduced. This metric is defined as the ratio between the total number of mode transitions and the overall interaction time, and it provides an indicator of the system’s robustness against unintended or spurious mode changes.

At present, there is no ISO/IEEE standard that formally defines quantitative performance ranges to guarantee fluid and natural interaction in gesture-based user interfaces. However, several studies in the field of Human--Computer Interaction and real-time gesture control provide empirical indications on acceptable levels of latency, stability, and control continuity required for a good user experience \cite{I29,I30,I31}. Based on these contributions, a fluid band was therefore defined, consisting of reference intervals for the main evaluation metrics: a mode switching rate between 2 and 5 switches per second, a command generation ratio between 0.8 and 1.0, a normalized RMS jitter between 0.005 and 0.03, an average processing latency between 10 and 50 ms (with an effective frame rate $\geq$ 20 fps), and a rendering latency between 10 and 25 ms. Although these intervals do not represent an official standard, they enable a direct quantitative comparison, allowing an objective assessment of the robustness and fluidity of the proposed system.

\section{RESULTS}
\begin{figure}
    \centering
    \includegraphics[width=1\linewidth]{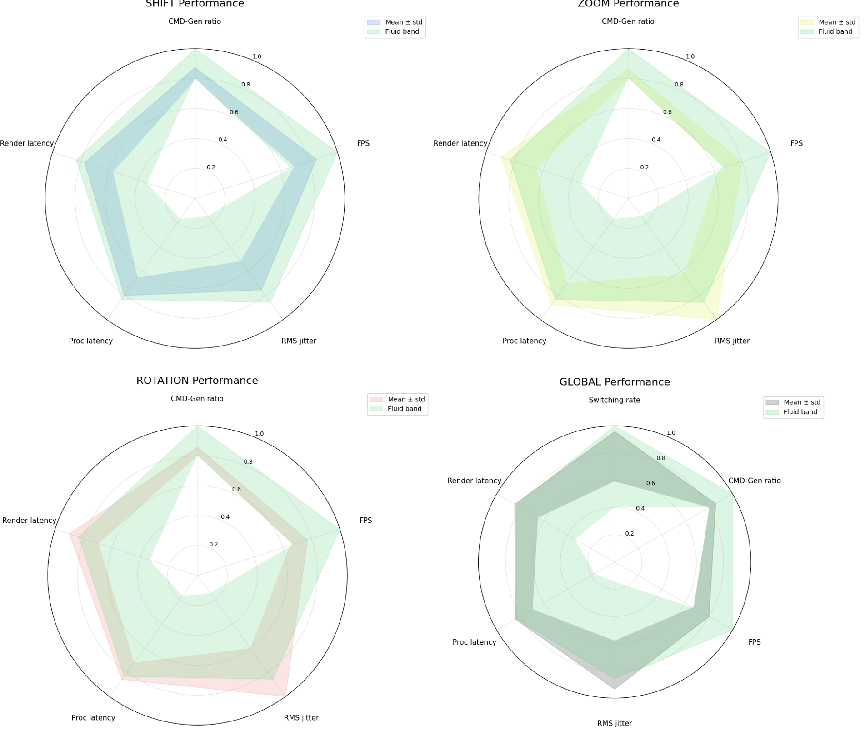}
    \caption{Global performance radar plot showing the mean normalized metrics, the $mean \pm standard$ deviation band, and the reference fluid band used to assess interaction robustness.}
    \label{comparison}
\end{figure}

After 10 experimental sessions, a total of 100,073 records were collected and analyzed. The results are first presented from a global perspective and subsequently detailed for each interaction mode.
From a global perspective, a total of 82,997 commands were generated across all interaction modes. The overall command-generation ratio was $0.83 \pm 0.03$, while the mode switching rate was measured at $3.87 \pm 0.92$ switches per second. The RMS jitter was equal to $2.60\% \pm 0.60\%$. The average processing latency was $45.99 \pm 4.47$~ms, and the average rendering latency was $22.44 \pm 3.00$~ms, corresponding to an estimated global frame rate of $21.94 \pm 2.08$~fps. For the SHIFT mode, 29,643 frames associated with active commands were recorded. The command-generation ratio was $0.83 \pm 0.04$. The RMS jitter was $2.20\% \pm 0.40\% $, while the average processing and rendering latencies were $43.51 \pm 4.58$~ms and $23.20 \pm 2.38$~ms, respectively. These values correspond to an estimated frame rate of $20.22 \pm 3.03$~fps. For the ROTATE mode, 34,557 action frames were collected, with a command-generation ratio of $0.83 \pm 0.03$. This mode exhibited a higher RMS jitter of $2.90\% \pm 0.80\%$. The average processing latency was $47.31 \pm 4.28$~ms, while the average rendering latency was $23.87 \pm 3.13$~ms, resulting in an estimated frame rate of $21.29 \pm 1.89$~fps. For the ZOOM mode, 18,797 action frames were recorded. The command-generation ratio was $0.83 \pm 0.04$. The RMS jitter was $2.90\% \pm 0.70\%$, while the average processing and rendering latencies were $47.18 \pm 5.41$~ms and $23.09 \pm 3.76$~ms, respectively. This mode achieved an estimated frame rate of $21.44 \pm 2.44$~fps. The normalized results are summarized graphically in Figure~\ref{comparison} and the quantitative ones are reported in the table \ref{results}.

\begin{table*}[ht]

\centering
\caption{System-level performance of the proposed touchless interface across individual gesture modes and overall system behavior}
\resizebox{\textwidth}{!}{
\begin{tabular}{cccccccccccccc}
\toprule
\multicolumn{1}{c}{} & \multicolumn{2}{c}{\textbf{CMD-Gen ratio}} & \multicolumn{2}{c}{\textbf{RMS jitter (\%)}} & \multicolumn{2}{c}{\textbf{Proc latency ($ms$)}}& \multicolumn{2}{c}{\textbf{Render latency ($ms$)}}& \multicolumn{2}{c}{\textbf{FPS}} & \multicolumn{2}{c}{\textbf{Switching rate}} \\
\cmidrule(rl){2-3} \cmidrule(rl){4-5}\cmidrule(rl){6-7}\cmidrule(rl){8-9}\cmidrule(rl){10-11} \cmidrule(rl){12-13}
\textbf{Mode} & {Mean $\uparrow$} & {Std Dev $\downarrow$} & {Mean $\downarrow$} & {Std Dev $\downarrow$}& {Mean $\downarrow$} & {Std Dev $\downarrow$}& {Mean $\downarrow$} & {Std Dev $\downarrow$}& {Mean $\uparrow$} & {Std Dev $\downarrow$} & {Mean} & {Std Dev} \\
\midrule
SHIFT & 0.83 & 0.04 & \textbf{2.20} & \textbf{0.40} & \textbf{43.51} & 4.58 & \textbf{20.22} & \textbf{3.03} & \textbf{23.20} & 2.38 \\
ROTATION & 0.83 & \textbf{0.03} & 2.90 & 0.80 & 47.31 & \textbf{4.28} & 23.87 & 3.13 & 21.29 & \textbf{1.89} \\
ZOOM & 0.83 & 0.04 & 2.90 & 0.70 & 47.18 & 5.41 & 23.09 & 3.76 & 21.44 & 2.44 \\
\midrule
GLOBAL & 0.83 & 0.03 & 2.60 & 0.60 & 45.99 & 4.47 & 22.44 & 3.00 & 21.94 & 2.08 & 3.87 & 0.92 \\
\bottomrule
\end{tabular}
}
\label{results}
\end{table*}

\section{DISCUSSION}
This work proposes an innovative approach for touchless control of medical image visualization systems in a surgical setting. Unlike existing solutions, which often require dedicated sensors, wearable devices, or active tracking techniques, the proposed system relies exclusively on a single RGB camera and is capable of capturing hand movements and mapping them directly into control commands for medical image navigation. In the present study, for implementation simplicity, the system was integrated with PyVista; however, the architecture is fully compatible with other visualization software widely used in the medical field, such as ImageJ and 3D Slicer, through the use of dedicated plugins. A further advantage of the system lies in its ease of use. The interaction is based on intuitive gestures, similar to those commonly used on touchscreen devices, and does not require long training periods, unlike other gesture-based systems proposed in the literature. Gesture initiation and termination are automatically managed through the depth information provided by MediaPipe Hands, enabling continuous and natural interaction. The fluidity and robustness of the control are confirmed by quantitative performance metrics, which show reduced latencies, signal stability, and behavior consistent with real-time interaction requirements.

Despite the encouraging results, this study presents some limitations. All test sessions were conducted in a controlled environment, with favorable lighting conditions. In a real surgical context, lighting is often reduced or non-uniform, which may affect the performance of the hand tracking model and, consequently, the fluidity of the interaction. In addition, factors such as partial hand occlusions, rapid movements, and sensor noise may introduce instability in gesture recognition, potentially leading to unintended commands or reduced tracking accuracy. Although not systematically evaluated in this study, these conditions represent critical scenarios for real-world deployment and should be addressed in future robustness analyses or stress-testing frameworks.
Another limitation concerns the latencies and the effective FPS, which are close to their minimum acceptable values. This leaves a limited operational margin and may affect the overall robustness of the system in stress cases. Potential mitigation actions should be considered to improve performance stability and resilience. Furthermore, usability evaluation was limited to technical tests: future studies involving surgeons and clinical staff will be necessary to collect qualitative and quantitative feedback on the actual usefulness of the system in the operating room. Additional limitations include the fact that the number of interaction modes currently supported is limited to translation, rotation, and zoom operations; the extension to more complex gestures and multi-hand scenarios represents a direction for future research.

\section{CONCLUSION}
This work presents a touchless system for intraoperative medical image interaction based on a single RGB camera and MediaPipe Hands. The proposed approach enables intuitive and real-time control without the need for dedicated hardware. Experimental results demonstrate its feasibility and robustness under controlled conditions. These findings support the potential of lightweight, vision-based interfaces for sterile surgical environments, motivating further validation in real clinical settings.

\section*{Acknowledgment}

This work was supported by the project Cal.Hub.Ria (project code T4-AN-09) funded by the Italian Ministry of Health in the framework of  "Piano Sviluppo e Coesione Salute, FSC 2014-2020".


\begin{thebibliography}{00}
\bibitem{I1}
Allegranzi, Benedetta, et al. "New WHO recommendations on preoperative measures for surgical site infection prevention: an evidence-based global perspective." The Lancet Infectious Diseases 16.12 (2016): e276-e287.

\bibitem{I2}
Weber, David J., Deverick Anderson, and William A. Rutala. "The role of the surface environment in healthcare-associated infections." Current opinion in infectious diseases 26.4 (2013): 338-344.

\bibitem{I3}
Cook, T. M., et al. "Consensus guidelines for managing the airway in patients with COVID‐19: Guidelines from the Difficult Airway Society, the Association of Anaesthetists the Intensive Care Society, the Faculty of Intensive Care Medicine and the Royal College of Anaesthetists." Anaesthesia 75.6 (2020): 785-799.

\bibitem{I4}
Dexter, Franklin, et al. "Perioperative COVID-19 defense: an evidence-based approach for optimization of infection control and operating room management." Anesthesia \& Analgesia 131.1 (2020): 37-42.

\bibitem{I5}
O’Hara, Kenton, et al. "Interactional order and constructed ways of seeing with touchless imaging systems in surgery." Computer Supported Cooperative Work (CSCW) 23.3 (2014): 299-337.

\bibitem{I6}
Mentis, Helena M., et al. "Interaction proxemics and image use in neurosurgery." Proceedings of the SIGCHI Conference on Human Factors in Computing Systems. 2012.

\bibitem{I7}
Wachs, Juan P., et al. "A gesture-based tool for sterile browsing of radiology images." Journal of the American Medical Informatics Association 15.3 (2008): 321-323.

\bibitem{I8}
Jacob, Mithun George, Juan Pablo Wachs, and Rebecca A. Packer. "Hand-gesture-based sterile interface for the operating room using contextual cues for the navigation of radiological images." Journal of the American Medical Informatics Association 20.e1 (2013): e183-e186.

\bibitem{I9}
Liu, Zhengnan, et al. "Advances in the development and application of non-contact intraoperative image access systems." BioMedical Engineering OnLine 23.1 (2024): 108.

\bibitem{I10}
Mewes, Andre, et al. "Touchless interaction with software in interventional radiology and surgery: a systematic literature review." International journal of computer assisted radiology and surgery 12.2 (2017): 291-305.

\bibitem{I11}
Besançon, Lonni, et al. "The state of the art of spatial interfaces for 3D visualization." Computer Graphics Forum. Vol. 40. No. 1. 2021.

\bibitem{I12}
Gallo, Luigi, Alessio Pierluigi Placitelli, and Mario Ciampi. "Controller-free exploration of medical image data: Experiencing the Kinect." 2011 24th international symposium on computer-based medical systems (CBMS). IEEE, 2011.

\bibitem{I13}
Madapana, Naveen, et al. "Touchless interfaces in the operating room: A study in gesture preferences." International Journal of Human–Computer Interaction 39.3 (2023): 438-448.

\bibitem{I14}
LIU, Jiaqing, et al. "A preliminary study of kinect-based real-time hand gesture interaction systems for touchless visualizations of hepatic structures in surgery." Medical Imaging and Information Sciences 36.3 (2019): 128-135.

\bibitem{I15}
Ebert, Lars C., et al. "You can’t touch this: touch-free navigation through radiological images." Surgical innovation 19.3 (2012): 301-307.

\bibitem{I16}
Rosa, Guillermo M., and María L. Elizondo. "Use of a gesture user interface as a touchless image navigation system in dental surgery: Case series report." Imaging science in dentistry 44.2 (2014): 155.

\bibitem{I16a}
Lin, Yin, et al. "Glioblastoma Overall Survival Prediction With Vision Transformers." 2025 47th Annual International Conference of the IEEE Engineering in Medicine and Biology Society (EMBC). IEEE, 2025.

\bibitem{I16b}
Lin, Yin, et al. "Lightweight ensemble vision transformer framework for non-invasive survival prediction in glioblastoma." Neurocomputing (2026): 133303.


\bibitem{I17}
Sa-nguannarm, Phataratah, et al. "A method of 3d hand movement recognition by a leap motion sensor for controlling medical image in an operating room." 2019 First International Symposium on Instrumentation, Control, Artificial Intelligence, and Robotics (ICA-SYMP). IEEE, 2019.

\bibitem{I18}
Feng, Yuanyuan, et al. "Comparison of kinect and leap motion for intraoperative image interaction." Surgical innovation 28.1 (2021): 33-40.

\bibitem{I19}
Kumar, Pradeep, et al. "A multimodal framework for sensor based sign language recognition." Neurocomputing 259 (2017): 21-38.

\bibitem{I20}
Kumar, Pradeep, et al. "Coupled HMM-based multi-sensor data fusion for sign language recognition." Pattern Recognition Letters 86 (2017): 1-8.

\bibitem{I21}
Mohamed, Noraini, Mumtaz Begum Mustafa, and Nazean Jomhari. "A review of the hand gesture recognition system: Current progress and future directions." IEEE access 9 (2021): 157422-157436.

\bibitem{I22}
Tompson, Jonathan, et al. "Real-time continuous pose recovery of human hands using convolutional networks." ACM Transactions on Graphics (ToG) 33.5 (2014): 1-10.

\bibitem{I23}
Oberweger, Markus, Paul Wohlhart, and Vincent Lepetit. "Hands deep in deep learning for hand pose estimation." arXiv preprint arXiv:1502.06807 (2015).

\bibitem{I24}
Simon, Tomas, et al. "Hand keypoint detection in single images using multiview bootstrapping." Proceedings of the IEEE conference on Computer Vision and Pattern Recognition. 2017.

\bibitem{I25}
Zimmermann, Christian, and Thomas Brox. "Learning to estimate 3d hand pose from single rgb images." Proceedings of the IEEE international conference on computer vision. 2017.

\bibitem{I26}
Molchanov, Pavlo, et al. "Hand gesture recognition with 3D convolutional neural networks." Proceedings of the IEEE conference on computer vision and pattern recognition workshops. 2015.

\bibitem{I27}
Zhang, Fan, et al. "Mediapipe hands: On-device real-time hand tracking." arXiv preprint arXiv:2006.10214 (2020).

\bibitem{I28}
Sullivan, C., and Alexander Kaszynski. "PyVista: 3D plotting and mesh analysis through a streamlined interface for the Visualization Toolkit (VTK)." Journal of Open Source Software 4.37 (2019): 1450.

\bibitem{I29}
Chen, Jessie YC, and Jennifer E. Thropp. "Review of low frame rate effects on human performance." IEEE Transactions on Systems, Man, and Cybernetics-Part A: Systems and Humans 37.6 (2007): 1063-1076.

\bibitem{I30}
Marvel, Jeremy A., et al. "Towards effective interface designs for collaborative HRI in manufacturing: metrics and measures." ACM Transactions on Human-Robot Interaction (THRI) 9.4 (2020): 1-55.

\bibitem{I31}
Niu, Peixin. "Convolutional neural network for gesture recognition human-computer interaction system design." PloS one 20.2 (2025): e0311941.

\end{thebibliography}
\end{document}